\pdfoutput=1
\documentclass[letterpaper, 10 pt, conference]{ieeeconf}  

\IEEEoverridecommandlockouts

\usepackage{subfigure}
\usepackage{graphics} 
\usepackage{epsfig} 
\usepackage{mathptmx} 
\usepackage{times} 
\usepackage{amsmath} 
\usepackage{amssymb}  
\usepackage{color}
\usepackage{xcolor}
\usepackage{multirow}

\usepackage{makecell}
\usepackage{cite}
\usepackage{hyperref}
\hypersetup{colorlinks}
\let\labelindent\relax
\usepackage{enumitem}
\usepackage{dsfont}
\usepackage{xurl}

\title{\LARGE \bf
Influence of Camera-LiDAR Configuration on 3D Object Detection for Autonomous Driving
}

\author{Ye Li$^{1*}$ \ Hanjiang Hu$^{2*}$ \ Zuxin Liu$^{2}$ \ Xiaohao Xu$^{1}$ \ Xiaonan Huang$^{1}$ \ Ding Zhao$^{2}$
\thanks{This work is supported by Office of Naval Research (Grant \#: N00014-24-1-2137; Program Manager: Michael “Q” Qin)}
\thanks{*The first two authors contributed equally}
\thanks{$^{1}$Ye Li, Xiaohao Xu, and Xiaonan Huang are with the Robotics Department, University of Michigan, Ann Arbor, MI 48109, USA. {\tt\small yeyli@umich.edu} }
\thanks{$^{2}$Hanjiang Hu is with the Machine Learning Department, and Zuxin Liu, Ding Zhao are with the Department of Mechanical Engineering, Carnegie Mellon University, Pittsburgh, PA 15213, USA. {\tt\small hanjianh@cs.cmu.edu} }
}

\begin{document}

\maketitle
\thispagestyle{empty}
\pagestyle{empty}

\begin{abstract}

Cameras and LiDARs are both important sensors for autonomous driving, playing critical roles in 3D object detection. Camera-LiDAR Fusion has been a prevalent solution for robust and accurate driving perception. In contrast to the vast majority of existing arts that focus on how to improve the performance of 3D target detection through cross-modal schemes, deep learning algorithms, and training tricks, we devote attention to the impact of sensor configurations on the performance of learning-based methods. To achieve this, we propose a unified information-theoretic surrogate metric for camera and LiDAR evaluation based on the proposed sensor perception model. We also design an accelerated high-quality framework for data acquisition, model training, and performance evaluation that functions with the CARLA simulator. To show the correlation between detection performance and our surrogate metrics, We conduct experiments using several camera-LiDAR placements and parameters inspired by self-driving companies and research institutions. Extensive experimental results of representative algorithms on nuScenes dataset validate the effectiveness of our surrogate metric, demonstrating that sensor configurations significantly impact point-cloud-image fusion based detection models, which contribute up to 30\% discrepancy in terms of the average precision.

\end{abstract}

\section{Introduction}

Multi-sensor fusion plays an important role in autonomous driving perception. Existing 3D object detection algorithms based on sensor fusion mainly use cameras and LiDAR. Cameras capture the rich texture and features of 3D objects in the environment \cite{he2017mask,yolo}, and LiDARs obtain the geometric characteristics of 3D objects \cite{wang2021object, LPD-Net,qiao2021registration}. In this paper, we study the perception system from the perspective of the physical design of multiple sensors, rather than fusion algorithms, and focus on the influence of camera-LiDAR configurations on 3D object detection performance.

\begin{figure}[ht]
\centering
\includegraphics[scale=0.38]{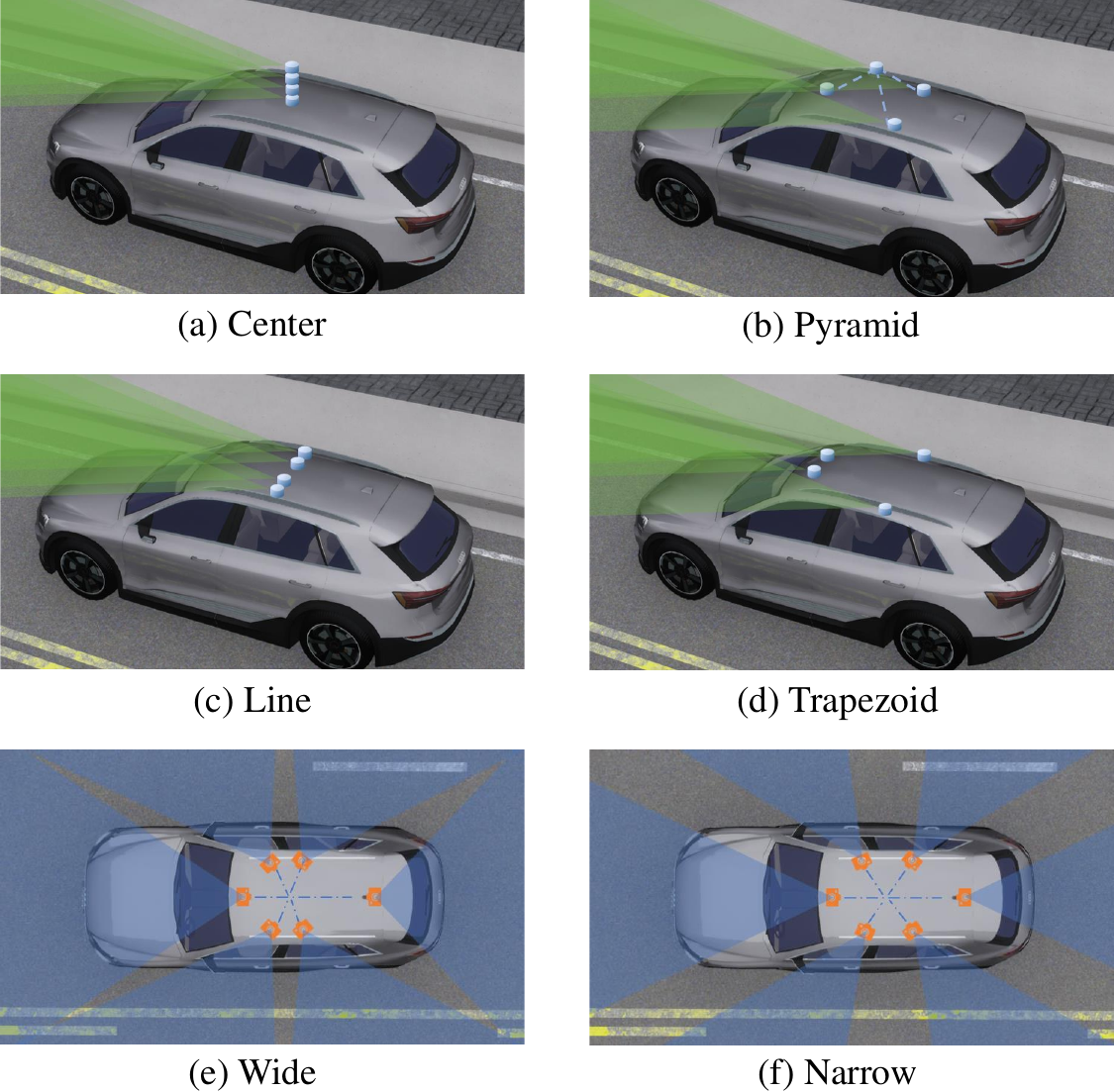}
\caption{LiDAR configurations of (a) Center, (b) Pyramid, (c) Line, (d) Trapezoid, and camera configurations of (e) Wide and (f) Narrow.}
\label{Audi}
\end{figure}

\begin{figure*}[ht]
\centering
\includegraphics[scale=0.4]{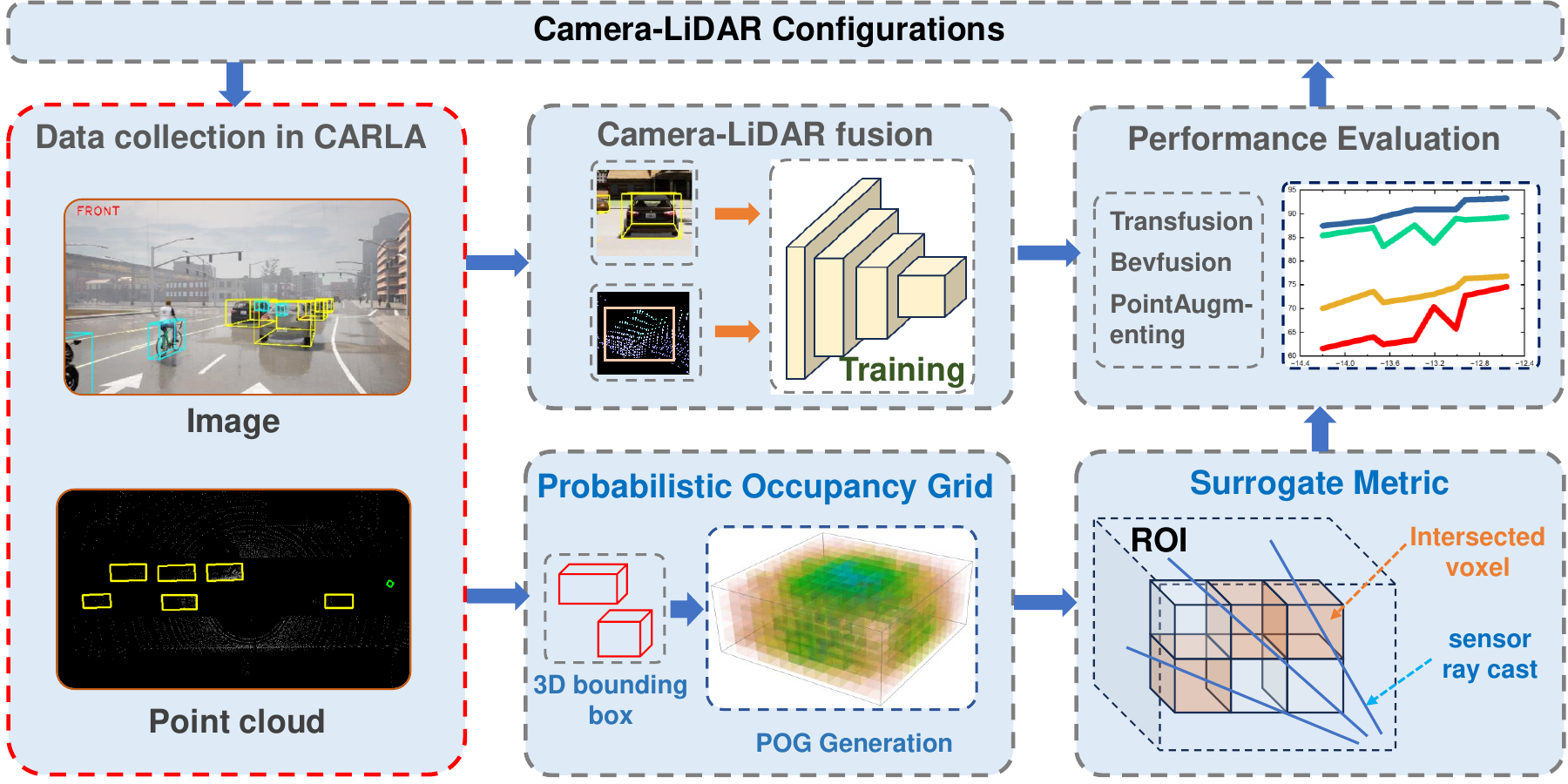}
\caption{Evaluation framework for camera-LiDAR configurations.}
\label{figure02}
\vspace{-8pt}
\end{figure*}

Well-designed sensor configurations are critical for autonomous driving, as improper camera and LiDAR configurations can lead to poor input data quality, which can affect detection performance  \cite{SPG}. Previous works  \cite{TransFusion, bevfusion, PointAugmenting} have explored a number of novel camera-LiDAR fusion perception algorithms, achieving excellent accuracy on autonomous driving datasets, such as nuScenes  \cite{nuscenes} and Waymo  \cite{waymo}. However, only a few preliminary arts  \cite{hu2022investigating, mou2021an, ma2021pe, ContinuousCamera} study the perception problem from the sensor-configuration perspective, e.g. different placements or parameters of sensors. Most current research focuses on  LiDAR  \cite{PlaceLiDARs, Twosensor} and camera  \cite{LargeCamera} configurations for sensing performance separately, but rarely establishes consistent criteria for both sensors. To this end, we aim to investigate the unified evaluation methods for both camera and LiDAR configurations, as shown in Fig. \ref{Audi}.

Fast evaluation of 3D detection performance under different camera and LiDAR configurations in the real world is quite challenging due to the laboriousness of data acquisition, model training, and performance testing  \cite{hu2022investigating}. Besides, efficiently comparing different sensor configurations for better 3D perception remains an open and critical question under a general trend of using more multi-modal sensors in autonomous driving \cite{wang}. To this point, this paper investigates the impact of camera-LiDAR configuration on 3D object detection performance and proposes a novel and unified framework for accelerating the evaluation of different camera-LiDAR configurations. The main contributions of this paper are summarized as follows:

\setlength{\itemsep}{0pt}
\setlength{\parsep}{0pt}
\setlength{\parskip}{0pt}
\begin{itemize}[leftmargin=*]
\item{We establish a new 
 systematic framework to efficiently evaluate the 3D detection performance of different camera-LiDAR configurations without the effort-costly loop of data collection, model training, and evaluation. }
\item{We propose an easy-to-compute unified surrogate metric based on the sensing mechanisms of both cameras and LiDARs, effectively characterizing the sensing procedure and accelerating the evaluation of perception performance. }
\item{
Experimental results in CARLA validate the correlation between our unified surrogate metric and the performance of several camera-LiDAR algorithms. The code is available on \href{https://github.com/ywyeli/lidar-camera-placement}{https://github.com/ywyeli/lidar-camera-placement}.}
\end{itemize}

\section{Related Work}

\subsection{Multi-modal Sensor Configurations}
Perception is an important sub-module of the autonomous driving system \cite{RichMaps}, which directly affects the decision-making and behavior of vehicles. With this respect, LiDARs and cameras are widely used for the perception of autonomous vehicles due to their capability to obtain rich information from the environment in real-time settings \cite{kitti, LOAM, OmniDet}. However, the performance of cameras and LiDARs is sensitive to physical installation errors or motion perturbations \cite{Twosensor,pmlr-v205-hu23b,hu2023robustness}.
There exist some works exploring configurations of cameras and LiDARs separately. Rahimian \textit{et al.}  \cite{CameraCapture} introduce a dynamic occlusion optimization method to estimate the configuration of the single camera. Puligandla \textit{et al.}  \cite{ContinuousCamera} optimize multi-camera placements for vehicle surround view in continuous domain using gradient-free black-box optimization. For LiDARs, Liu \textit{et al.}  \cite{Liuzuxin} develop a non-detectable-space-based surrogate function and realize the optimum with the Artificial Bee Colony algorithm. 
Instead of optimizing the surrounding view of sensors on the vehicle to minimize undetectable space, recent works  \cite{hu2022investigating} quantitatively investigate the interaction between LiDAR placement and detection performance and propose metrics to evaluate LiDAR perception. However, quantitative evaluation methods for both cameras and LiDARs associated with sensor fusion based 3D detection algorithms remain to be explored although \cite{ma2021pe} initiates the study of multiple sensors empirically. To this end, we present a novel unified metric for quantitatively studying both camera and LiDAR configurations in Section \ref{sec:method}.


\subsection{Camera-LiDAR Detection Algorithms}

3D object detection is an important task in autonomous driving. Recently, camera-LiDAR fusion has attracted increasing attention to boost more robust and accurate detection performance. Current camera-LiDAR fusion work can be classified into three categories: result-level, proposal-level, and point-level. The result-level methods  \cite{FPointNet,FConvNet, RoarNet, PointNet} use 2D object detectors to prepare 3D detection proposals, with difficulty encountered when objects overlap in 2D planes. The proposal-level methods  \cite{AVOD, MV3D} generate 3D proposals directly and perform fusion at the region-of-interest (ROI) level through ROIPool  \cite{FasterR-CNNRoIPool}, but the granularity is poor. Most recent works use the point-level method and achieve promising detection performance. Point-level fusion methods established the association between 3D points and image pixels through calibration matrices and decorate LiDAR points input with image segmentation features, including input-level decoration  \cite{Pointpainting, PointAugmenting, MVP, FusionPainting, AutoAlign, FocalSparseCNN} and feature-level decoration. Moreover, some approaches  \cite{Multi-TaskMulti-SensorFusion, Pi-rcnn, 3D-CVF} perform the decoration process on the bird's eye view (BEV) plane. 
The majority of the detection methods mentioned above are well-designed and evaluated on high-quality point cloud datasets, but they do not consider the placements of the camera-LiDAR sensing system. We adopt several representative latest methods to evaluate the influence of camera-LiDAR configurations on detection performance in Section \ref{sec:exp}.

\section{Unified Surrogate Metric}
\label{sec:method}
We propose a new surrogate metric derived from maximal information gain, leveraging camera and LiDAR ray-casting algorithms. First, we compute the probabilistic occupancy grid and the entropy of the joint distribution for scenes equipped with bounding boxes ground truth. Then we identified the voxels intersected by rays from both cameras and LiDARs to determine the conditional entropy and the unified surrogate metric of information gain, as shown in Fig. \ref{figure01}. 

\begin{figure}[ht]
\centering
\includegraphics[scale=0.45]{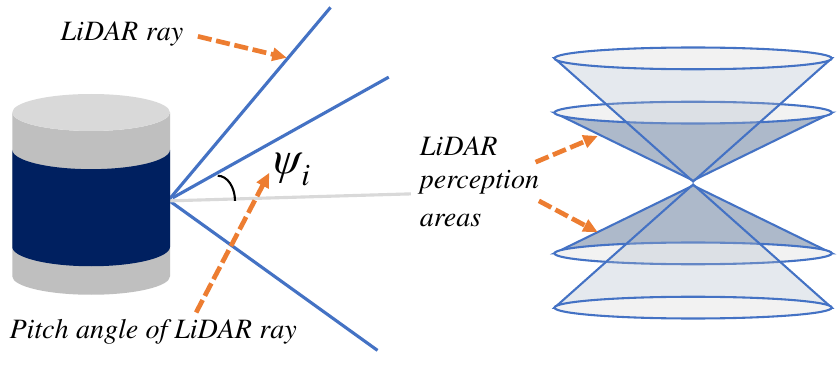}
\caption{LiDAR sensing model}
\label{lidar_model}
\end{figure}
\begin{figure}[htbp]
\centering
\includegraphics[scale=0.4]{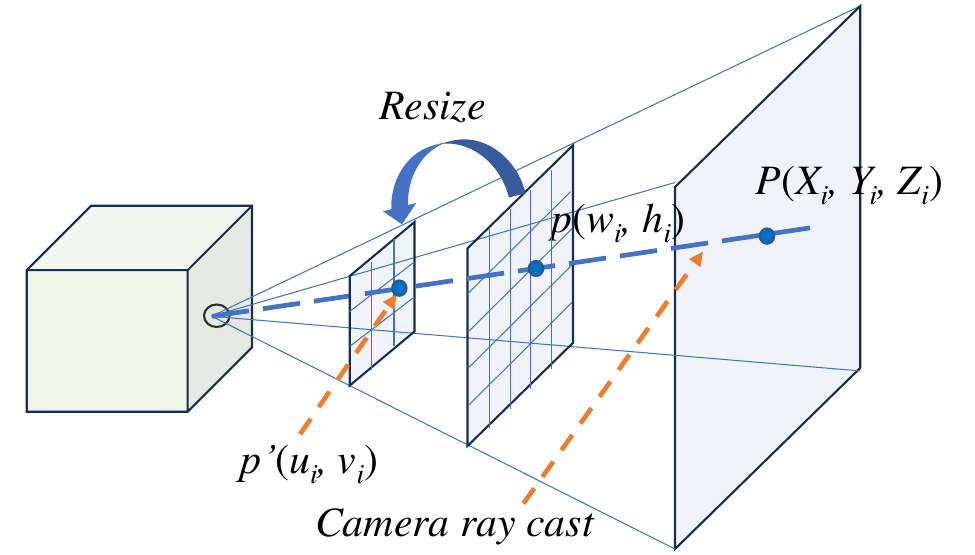}
\caption{Camera sensing model}
\label{camera_model}
\end{figure}


\subsection{Problem Formulation}

To evaluate the performance of different camera-LiDAR configurations, we only consider the objects in the region of interest (ROI) when calculating the detection accuracy metrics. We formulate the problem of camera-LiDAR configuration evaluation as comparing the 3D object detection performance with several state-of-the-art camera-LiDAR fusion methods and their corresponding surrogate metrics. Given the difficulty of evaluating the performance of camera-LiDAR detection in the real world, we propose a unified surrogate metric to accelerate the sensor configuration evaluation procedure. We introduce the camera-LiDAR perception model to calculate the unified surrogate metric.

In our work, ROI refers to the limited 3D cuboid area $[L, W, H]$, where the sensor detects the objects. We divide the ROI area $\Omega$ into voxels of the same resolution $\delta$,
\begin{equation}
\Omega = \{\omega_1,\omega_2,...,\omega_N\}, N = \frac{L}{\delta}\times\frac{W}{\delta}\times\frac{H}{\delta}
\end{equation}
where $N$ denotes the total number of divided voxels $\omega_i (i=1,2,...,N)$ in the ROI. 

\subsection{Modeling Camera-LiDAR Perception}

We first introduce the camera-LiDAR data sensing model. To evaluate the parameters of LiDAR and camera with a unified method, we propose the following ray cast method of both camera and LiDAR. 

\textbf{LiDAR perception model.} Inspired by mechanically rotating LiDAR, we model the LiDAR perception model as a series of rotating vertically distributed rays. These rays are evenly distributed at certain angular intervals in the vertical direction, and they rotate synchronously around the vertical Z axis at a fixed angular velocity. Each ray has a fixed pitch angle to the horizontal plane and forms a conical surface by rotation. The sensing area of LiDAR is the collection in the ROI of all conical surfaces. For a LiDAR of ${horizontal = \theta^L_0, vertical = \psi^L_0}$, we calculate the yaw $\theta^L_{ij}$ and the pitch $\psi^L_{ij}$ rotation angles for these rays as,
\begin{equation}
\theta^L_{ij} = \frac{\theta^L_0}{i}, 
\psi^L_{ij} = \frac{\psi^L_0}{j} + \frac{\psi^L_0}{2},
\quad i \in \{1,...,I\},j \in \{1,...,J\}
\end{equation}
where $I$ and $J$ represent the horizontal and vertical resolution for the specific LiDAR.

\textbf{Camera perception model.} Based on the pin-hole image camera model with image projection \cite{hartley2003multiple}, as shown in Fig. \ref{camera_model}, we can obtain the transformation matrix between the coordinates of each pixel and the points in the 3D world. Each pixel point $p(w_i,h_i)$ and the corresponding 3D world point $P(X_i, Y_i, Z_i)$ form a ray with its origin at the camera's optical center $O$. The RGB color channel of each pixel represents the exterior features of the object encountered by the ray.  
Thus, the total perception of a camera is the union of all ROI space traversed by the rays. Therefore, we see the fusion sensing model of LiDAR and camera as two sets of rays scanned in the ROI. 

In our camera perception model, the number of rays does not exactly equal the original pixel resolution $w_0 \times h_0$
. The images obtained by the camera are dense and with high resolution, but learning-based methods do not directly take the original images as inputs without down-sampling. So the original images are resized to a lower resolution $u_0 \times v_0$
to model the process of the camera ray casting. We first map the pixels on the resized image to the pixels on the original image as follows,
\begin{equation}
w_i = u_i \times \frac{w_0}{u_0}, h_i = v_i \times \frac{h_0}{v_0}, \quad
i=1,2,...,u_0 \times v_0
\end{equation}

Given pixel focal length of the camera $f$, we calculate the yaw $\theta^c_i$ and pitch $\psi^c_i$ rotation angles for each ray $i=1,2,...,u_0 \times v_0$, 
\begin{align}
    \theta^c_i &= acrcos\frac{w_i-\frac{w_0}{2}}{\sqrt{(w_i-\frac{w_0}{2})^2+f^2}} \\
    \psi^c_i &= acrcos\frac{h_i-\frac{h_0}{2}}{\sqrt{(w_i-\frac{w_0}{2})^2+(h_i-\frac{h_0}{2})^2+f^2}}
\end{align}


\begin{table*}[ht]
\caption{Unified Surrogate Metric for Multi-sensor Configuration}
\label{tab1}
\center
\begin{tabular}{|c||ccc||ccc||ccc|}
\hline
\multirow{2}{*}{Sensor Configurations} & \multicolumn{3}{c||}{Camera S-MIG ($10^3$)}                                 & \multicolumn{3}{c||}{LiDAR S-MIG ($10^3$)}                                 & \multicolumn{3}{c|}{$\lambda=0.1$, S-MS ($10^3$)}                                         \\ \cline{2-10} 
                                       & \multicolumn{1}{c|}{Car}    & \multicolumn{1}{c|}{Bicycle} & Pedestrian & \multicolumn{1}{c|}{Car}   & \multicolumn{1}{c|}{Bicycle} & Pedestrian & \multicolumn{1}{c|}{Car}    & \multicolumn{1}{c|}{Bicycle} & Pedestrian \\ \hline
Wide + Center (W+C)                       & \multicolumn{1}{c|}{-73.08} & \multicolumn{1}{c|}{-14.15}  & -13.13     & \multicolumn{1}{c|}{-6.45} & \multicolumn{1}{c|}{-1.07}   & -1.08      & \multicolumn{1}{c|}{-13.75} & \multicolumn{1}{c|}{-2.48}   & -2.39      \\ \hline
Wide + Pyramid (W+P)                         & \multicolumn{1}{c|}{-73.08} & \multicolumn{1}{c|}{-14.15}  & -13.13     & \multicolumn{1}{c|}{-5.90} & \multicolumn{1}{c|}{-0.96}   & -0.92      & \multicolumn{1}{c|}{-13.21} & \multicolumn{1}{c|}{-2.38}   & -2.23      \\ \hline
Wide + Line  (W+L)                       & \multicolumn{1}{c|}{-73.08} & \multicolumn{1}{c|}{-14.15}  & -13.13     & \multicolumn{1}{c|}{-5.62} & \multicolumn{1}{c|}{-0.91}   & -0.88      & \multicolumn{1}{c|}{-12.93} & \multicolumn{1}{c|}{-2.32}   & -2.19      \\ \hline
Wide + Trapezoid  (W+T)                         & \multicolumn{1}{c|}{-73.08} & \multicolumn{1}{c|}{-14.15}  & -13.13     & \multicolumn{1}{c|}{-5.25} & \multicolumn{1}{c|}{-0.87}   & -0.83      & \multicolumn{1}{c|}{-12.56} & \multicolumn{1}{c|}{-2.29}   & -2.14      \\ \hline
Narrow + Center   (N+C)                  & \multicolumn{1}{c|}{-77.56} & \multicolumn{1}{c|}{-15.57}  & -14.09     & \multicolumn{1}{c|}{-6.45} & \multicolumn{1}{c|}{-1.07}   & -1.08      & \multicolumn{1}{c|}{-14.20} & \multicolumn{1}{c|}{-2.62}   & -2.49      \\ \hline
Narrow + Pyramid   (N+P)                    & \multicolumn{1}{c|}{-77.56} & \multicolumn{1}{c|}{-15.57}  & -14.09     & \multicolumn{1}{c|}{-5.90} & \multicolumn{1}{c|}{-0.96}   & -0.92      & \multicolumn{1}{c|}{-13.66} & \multicolumn{1}{c|}{-2.52}   & -2.33      \\ \hline
Narrow + Line  (N+L)                   & \multicolumn{1}{c|}{-77.56} & \multicolumn{1}{c|}{-15.57}  & -14.09     & \multicolumn{1}{c|}{-5.62} & \multicolumn{1}{c|}{-0.91}   & -0.88      & \multicolumn{1}{c|}{-13.38} & \multicolumn{1}{c|}{-2.46}   & -2.29      \\ \hline
Narrow + Trapezoid   (N+T)                    & \multicolumn{1}{c|}{-77.56} & \multicolumn{1}{c|}{-15.57}  & -14.09     & \multicolumn{1}{c|}{-5.25} & \multicolumn{1}{c|}{-0.87}   & -0.83      & \multicolumn{1}{c|}{-13.01} & \multicolumn{1}{c|}{-2.43}   & -2.24      \\ \hline
\end{tabular}
\end{table*}

\subsection{Multi-sensor Probabilistic Occupancy Grid}

Following \cite{hu2022investigating}, it is assumed that the more 3D objects are covered by the rays of our camera-LiDAR seeing model, the better the performance of multi-sensor 3D detection will be. Based on this intuition, we first propose Probabilistic Occupancy Grid (POG) to evaluate the probability that each voxel in ROI is occupied by the target object to be detected.
\begin{equation}
p_{POG}= p\{\omega_1,\omega_2,...,\omega_N\}
\end{equation}
where $\omega_i \sim p_\Omega$ and $N$ is the number of voxels in ROI. Given a dataset with ground true of samples $Y_T = \{y_1,y_2,...,y_T\}$, $T$ denotes the total frames of ground truth samples. Each frame $y_t \in \{y_1,...,y_T\}$ of the given samples contains $S^{(t)}$ ground-truth bounding boxes. We denote the bounding boxes of target objects as $\{b_1^{(t)},b_2^{(t)},...,b_{S^{(t)}}^{(t)}\}$. We use $\omega_i \in y_t$ to denote the case where the voxel $\omega_i$ is contained by the ground truth bounding box of any target object in the frame $y_t$. Then, for each voxel $\omega_i$ in the ROI, we traverse all the $T$ frames of samples to obtain the probability that the voxel $\omega_i$ is occupied by the target objects in any given frame from $Y_T$ as,
\begin{equation}
\begin{aligned}
&\hat{p}(\omega_i) = \frac{\sum_{t=1}^T\mathds{1}(\omega_i \in y_t)}{T}, \\
&\omega_i \in y_t := \exists \; b^{(t)} \in y_t, \; s.t. \; \omega_i \in b^{(t)}
\end{aligned}
\end{equation}
where $\mathds{1}(\cdot)$ is an indicator function. The POG can then be estimated by the joint probability of all occupied voxels in ROI. Since the presence of an object in one voxel does not imply presence in other voxels among all the frames in ROI, we could treat these voxels as independent and identically distributed random variables and calculate the joint distribution over all non-zero voxels in the set $\Omega$ as,
\begin{equation}
\hat{p}_{POG} = \hat{p}(\omega_1,\omega_2,...,\omega_{N}) = \prod_{i=1,\hat{p}(\omega_i)\neq 0}^{N}{\hat{p}(\omega_i)}
\end{equation}
where $N$ is the total number of voxels in ROI. 
Note that notations of $\hat{p}$ with  $\hat{}$ are the estimated distribution from observed samples, while notations of $p$ without $hat$ are the unknown non-random statistics to be estimated.

Considering the ray casting sensing model for a specific multi-sensor configuration, we define the conditional Probabilistic Occupancy Grid to represent the conditional probability that a voxel is occupied by the 3D bounding boxes of the target objects in the perceptual field of the sensor, with the assumption of conditional independence. 
\begin{equation}
p_{POG|C=C_0}= p\{\omega_1,\omega_2,...,\omega_N \; | \; C=C_0\}
\end{equation}

To make the notation compact and easy to read, we denote the occupied voxel random variable $\omega_i | C=C_0$ as $\omega_i^{C_0}$ and denote the conditional distribution as $p_{\Omega | C=C_0}$ as $p_{\Omega | C_0}$, so we have $\omega_i \sim p_{\Omega | C_0}$.

We use Bresenham’s Line Algorithm  \cite{IBM} to deal with our proposed sensor ray casting sensing model, finding all voxels traversed by the ray generated from the given camera or LiDAR sensor.
For the specific sensor configuration $C = C_0$, we denote the set of voxels traversed by rays as,
\begin{equation}
\Omega| C_0 = Brsenham(\Omega,C_0) = \{\omega_1^{C_0},\omega_2^{C_0},...,\omega^{C_0}_N\}
\end{equation}

Given the specific sensor configuration $C = C_0$, the conditional Probabilistic Occupancy Grid can be expressed as,
\begin{equation}
\hat{p}_{POG|C_0} = \hat{p}(\omega_1^{C_0},\omega_2^{C_0},...,\omega^{C_0}_N) = \prod_{i=1,\hat{p}(\omega_i^{C_0})\neq0}^{N}{\hat{p}(\omega_i^{C_0})}
\end{equation}

The conditional POG reflects the conditional joint distribution of voxels that intersect with the perceptual area of the sensor in the ROI. From the perspective of density estimation to find POG, the true POG and conditional POG given configuration $C_0$ can be estimated as
\begin{equation}
p_{POG} = \hat{p}_{POG}, \\\ p_{POG|C_0} = \hat{p}_{POG|C_0}
\end{equation}


\subsection{Unified Surrogate Metric for Multi-sensor Fusion}

In this section, we propose a unified surrogate metric for evaluating camera-LiDAR configurations, based on POG and information theory. We first define the surrogate metric for a single type of sensor. Given specific sensor configurations, We denote the total uncertainty in the joint voxel distribution by the total entropy of POG $H_{POG}$. Further, we can represent the conditional uncertainty with the entropy of the conditional POG given sensor configurations $C=C_0$.
\begin{align}
    H_{POG} &= H(\Omega)=\mathds{E}_{\omega_i \sim p_\Omega} \sum_{i=1}^{N}\hat{H}(\omega_i) \\
    H_{POG|C_0} &= H(\Omega|C_0)=\mathds{E}_{\omega_i \sim p_\Omega} \sum_{i=1}^{N}\hat{H}(\omega_i^{C_0})
\end{align}

Perception performance of a given configuration is maximized when the uncertainty of the voxel distribution to the sensor is minimized\cite{hu2022investigating}. We denote the information gain $IG$ using the mutual information (\textbf{MI}) between the full entropy and the conditional entropy, which represents the decrease of the uncertainty of the voxel distribution given a specific sensor configuration.
\begin{equation}
IG_{\Omega,C_0} = H(\Omega) -H(\Omega|C_0) 
\end{equation}

Since the total entropy $H(\Omega)$ is a constant given POG with a fixed voxel distribution and irrelevant to different sensor configurations, we denote our maximum-information-gain-based surrogate metric S-MIG as,
\begin{equation}
S_{MIG\mid C_0} = -H(\Omega|C_0) = -\mathds{E}_{\omega_i \sim p_\Omega} \sum_{i=1}^{N}\hat{H}(\omega_i^{C_0})
\end{equation}

Further, we introduce the Unified Surrogate Metric for Multi-sensor Fusion to evaluate the multi-sensor configuration. For camera configuration $C^c(\theta^c,\psi^c)$ and LiDAR configuration $C^L(\theta^L,\psi^L)$, we denote Unified Surrogate Metric for Multi-sensor (\textbf{S-MS}) as,
\begin{equation}
S_{MS} = \lambda S_{MIG|C^c(\theta^c,\psi^c)} + S_{MIG|C^L(\theta^L,\psi^L)}
\end{equation}
where $\lambda$ is a hyper-parameter weighting the difference in perceptual capability between LiDAR and the camera, to fairly show the influence of cameras and LiDARs on 3D detection. 

\begin{table*}[htp]
\caption{Comparison of object detection performance under various camera-LiDAR configurations using different methods}
\center
\label{table01}
\begin{tabular}{|cc||cccc|cccc|}
\hline
\multicolumn{2}{|c||}{Car mAP}        & \multicolumn{4}{c|}{Wide} & \multicolumn{4}{c|}{Narrow} \\ \hline
\multicolumn{1}{|c|}{Model}           & Backbone       & \multicolumn{1}{c|}{Center} & \multicolumn{1}{c|}{Pyramid} & \multicolumn{1}{c|}{Line}             & Trapezoid      & \multicolumn{1}{c|}{Center} & \multicolumn{1}{c|}{Pyramid} & \multicolumn{1}{c|}{Line}             & Trapezoid        \\ \hline
\multicolumn{1}{|c|}{Transfusion}     & PointPillars   & \multicolumn{1}{c|}{87.08}  & \multicolumn{1}{c|}{83.87}   & \multicolumn{1}{c|}{89.08}            & \textbf{89.27} & \multicolumn{1}{c|}{85.44}  & \multicolumn{1}{c|}{83.1}    & \multicolumn{1}{c|}{87.62}            & \textbf{88.74}   \\ \hline
\multicolumn{1}{|c|}{Bevfusion}       & VoxelNet       & \multicolumn{1}{c|}{88.65}  & \multicolumn{1}{c|}{90.93}   & \multicolumn{1}{c|}{92.95}            & \textbf{93.29} & \multicolumn{1}{c|}{87.49}  & \multicolumn{1}{c|}{89.42}   & \multicolumn{1}{c|}{\textbf{90.96}}   & 90.91            \\ \hline
\multicolumn{1}{|c|}{PointAugmenting} & PointPillars   & \multicolumn{1}{c|}{73.68}  & \multicolumn{1}{c|}{73.07}   & \multicolumn{1}{c|}{76.35}            & \textbf{76.81} & \multicolumn{1}{c|}{70.04}  & \multicolumn{1}{c|}{71.36}   & \multicolumn{1}{c|}{72.35}            & \textbf{74.56}   \\ \hline
\multicolumn{1}{|c|}{PointAugmenting} & VoxelNet       & \multicolumn{1}{c|}{64.06}  & \multicolumn{1}{c|}{70.37}   & \multicolumn{1}{c|}{72.78}            & \textbf{74.62} & \multicolumn{1}{c|}{61.62}  & \multicolumn{1}{c|}{62.48}   & \multicolumn{1}{c|}{63.48}            & \textbf{67.8}    \\ \hline
\end{tabular}

\vspace{-4pt} 
\center
\begin{tabular}{|cc||cccc|cccc|}
\hline
\multicolumn{2}{|c||}{Bicycle mAP}  & \multicolumn{4}{c|}{Wide} & \multicolumn{4}{c|}{Narrow} \\ \hline
\multicolumn{1}{|c|}{Model}         & Backbone       & \multicolumn{1}{c|}{Center} & \multicolumn{1}{c|}{Pyramid} & \multicolumn{1}{c|}{Line}           & Trapezoid      & \multicolumn{1}{c|}{Center} & \multicolumn{1}{c|}{Pyramid} & \multicolumn{1}{c|}{Line}           & Trapezoid      \\ \hline
\multicolumn{1}{|c|}{Transfusion}     & PointPillars & \multicolumn{1}{c|}{78.45}  & \multicolumn{1}{c|}{80.63}   & \multicolumn{1}{c|}{83.98}          & \textbf{86.21} & \multicolumn{1}{c|}{72.79}  & \multicolumn{1}{c|}{77.87}   & \multicolumn{1}{c|}{79.04}          & \textbf{83.38} \\ \hline
\multicolumn{1}{|c|}{Bevfusion}       & VoxelNet     & \multicolumn{1}{c|}{83.62}  & \multicolumn{1}{c|}{89.55}   & \multicolumn{1}{c|}{90.53}          & \textbf{90.78} & \multicolumn{1}{c|}{79.95}  & \multicolumn{1}{c|}{83.89}   & \multicolumn{1}{c|}{86.22}          & \textbf{87.22} \\ \hline
\multicolumn{1}{|c|}{PointAugmenting} & PointPillars & \multicolumn{1}{c|}{64.88}  & \multicolumn{1}{c|}{67.17}   & \multicolumn{1}{c|}{\textbf{69.79}} & 68.98          & \multicolumn{1}{c|}{53.97}  & \multicolumn{1}{c|}{61.66}   & \multicolumn{1}{c|}{\textbf{65.26}} & 63.04          \\ \hline
\multicolumn{1}{|c|}{PointAugmenting} & VoxelNet     & \multicolumn{1}{c|}{56.38}  & \multicolumn{1}{c|}{62.18}   & \multicolumn{1}{c|}{\textbf{65.35}} & 63.94          & \multicolumn{1}{c|}{51.59}  & \multicolumn{1}{c|}{55.09}   & \multicolumn{1}{c|}{\textbf{57.46}} & 56.53          \\ \hline
\end{tabular}

\vspace{-4pt} 
\center
\begin{tabular}{|cc||cccc|cccc|}
\hline
\multicolumn{2}{|c||}{Pedestrian mAP} & \multicolumn{4}{c|}{Wide} & \multicolumn{4}{c|}{Narrow} \\ \hline
\multicolumn{1}{|c|}{Model}           & Backbone     & \multicolumn{1}{c|}{Center}   & \multicolumn{1}{c|}{Pyramid} & \multicolumn{1}{c|}{Line}           & Trapezoid      & \multicolumn{1}{c|}{Center}   & \multicolumn{1}{c|}{Pyramid}  & \multicolumn{1}{c|}{Line}           & Trapezoid      \\ \hline
\multicolumn{1}{|c|}{Transfusion}     & PointPillars & \multicolumn{1}{c|}{45.77}    & \multicolumn{1}{c|}{57.51}   & \multicolumn{1}{c|}{66.55}          & \textbf{68.06} & \multicolumn{1}{c|}{39.38}    & \multicolumn{1}{c|}{53.39}    & \multicolumn{1}{c|}{60.74}          & \textbf{65.97} \\ \hline
\multicolumn{1}{|c|}{Bevfusion}       & VoxelNet     & \multicolumn{1}{c|}{57.37}    & \multicolumn{1}{c|}{65.96}   & \multicolumn{1}{c|}{\textbf{72.22}} & 67.89          & \multicolumn{1}{c|}{55.39}    & \multicolumn{1}{c|}{58.65}    & \multicolumn{1}{c|}{\textbf{62.36}} & 62.21          \\ \hline
\multicolumn{1}{|c|}{PointAugmenting} & PointPillars & \multicolumn{1}{c|}{35.71}    & \multicolumn{1}{c|}{37.71}   & \multicolumn{1}{c|}{39.96}          & \textbf{43.77} & \multicolumn{1}{c|}{25.1}     & \multicolumn{1}{c|}{30.38}    & \multicolumn{1}{c|}{33.7}           & \textbf{38.14} \\ \hline
\multicolumn{1}{|c|}{PointAugmenting} & VoxelNet     & \multicolumn{1}{c|}{29.67}    & \multicolumn{1}{c|}{30.48}   & \multicolumn{1}{c|}{31.27}          & \textbf{32.36} & \multicolumn{1}{c|}{25.63}    & \multicolumn{1}{c|}{27.92}    & \multicolumn{1}{c|}{26.81}          & \textbf{29.63} \\ \hline
\end{tabular}
\end{table*}
\vspace{-4pt}
\section{Experiments}
\label{sec:exp}
In this section, we design a framework for automatic camera-LiDAR data collection and evaluation based on the CARLA simulator \cite{dosovitskiy2017carla} to avoid time-consuming, low-efficiency, and high-cost real-world experiments to validate our method. To ensure the fairness of the experimental comparison, all images and point clouds collected in CARLA are with fixed scenarios, including the route of the data-collection ego vehicle, driving scenarios, traffic flow, etc. We conducted comprehensive experiments to show two key points: the impact of camera and LiDAR configurations on 3D object detection performance, and the correlation between our proposed unified surrogate metric and learning-based perception performance.


\subsection{Experimental Setup}
We collect data in the CARLA simulator and calculate the S-MS for each camera-LiDAR configuration based on the bounding box of the 3D target objects.

\textbf{CARLA simulator and dataset.} In order to verify our proposed method, we design an automatic camera-LiDAR data collection pipeline based on the realistic CARLA simulator. Following \cite{hu2022investigating}, we collect data in Town 1, 3, 4, and 6 and each town contains 8 manually recorded routes. Only camera and LiDAR configurations are changed for all the experiments. We collected 8 nuScenes-like datasets with different camera-LiDAR configurations, and each dataset contains about 38000 frames. Each frame of the dataset contains six images and one set of point clouds with 3D bounding boxes of Car, Bicycle, and Pedestrian. We use the CARLA simulator to collect point cloud and image data in the nuScenes format \cite{caesar2020nuscenes}. We use 72\% frames as the training set, 18\% frames as the validation set, and the remaining 10\% frames as the test set.

\textbf{Camera-LiDAR fusion detection algorithms and metrics.} To fairly demonstrate the performance under different camera-LiDAR configurations on 3D object detection, we adopt the three representative open-source camera-LiDAR fusion detection methods. Specifically, we use two two-stage detection methods, Transfusion \cite{TransFusion} and Bevfusion \cite{bevfusion}, and one single-stage detection method, Pointaugmenting \cite{PointAugmenting}.  We follow the default training pipeline and hyperparameters of these methods. For two-stage methods, We first train the LiDAR-only detection model, then fine-tune the fusion methods with pretrained image detection models. For the one-stage method, we train two backbones, VoxelNet \cite{VoxelNet}, and Pointpillars \cite{PointPillars}, respectively. We adopt the mean Average Precision (mAP) metric from nuScenes \cite{caesar2020nuscenes}, which defines thresholds by considering the 2D distance on the ground rather than Intersection over Union (IoU) based ones \cite{kitti}.

\textbf{Different sensor configurations.} The sensor configurations in this work adopt 4 LiDARs and 6 cameras following nuScenes \cite{nuscenes2019}. Four different LiDAR configurations inspired by famous autonomous driving companies \cite{hu2022investigating} and two camera configurations are used to explore the influence on object detection performance. 
For LiDAR configurations, the beams are equally distributed in the vertical FOV [-25.0, 25.0] degrees, as shown in Fig. \ref{Audi}. Center placement is achieved by vertically stacking four LiDARs together at the roof center (\ref{Audi}a). Pyramid placement (Fig. \ref{Audi}b) includes 1 front LiDAR and 3 back ones, with a higher one in the middle. Line has (\ref{Audi}c) 4 LiDARs placed in a horizontal line symmetrically. Trapezoid installs 4 LiDARs in the parallel front and back (Fig. \ref{Audi}d). Each camera has the same resolution 1600×900. Wide configuration is the same as the setup of nuScenes with 5 general cameras (FOV = 70°) and a wide-angle camera (FOV = 110°) on the roof of the ego car (Fig. \ref{Audi}e). As contrast to the Wide configuration, we design another Narrow configuration by changing the FOV to 55 degrees for all cameras in Fig. \ref{Audi}f. 


\subsection{Experiment Results and Analysis}
We first  show the influence of camera-LiDAR configurations on 3D object detection average precision, and then analyze the relation between the detection performance and our proposed unified surrogate metric.

\begin{figure*}[ht]
\centering{
\begin{minipage}{2.3in}
\includegraphics[scale=0.75]{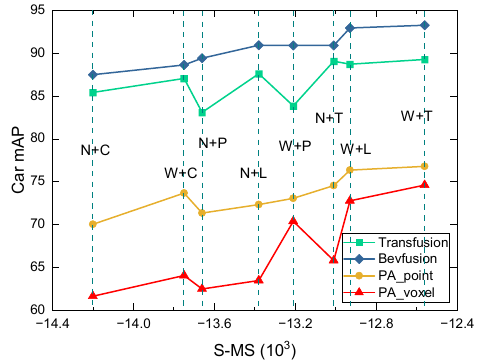}
\end{minipage}
\begin{minipage}{2.3in}
\includegraphics[scale=0.75]{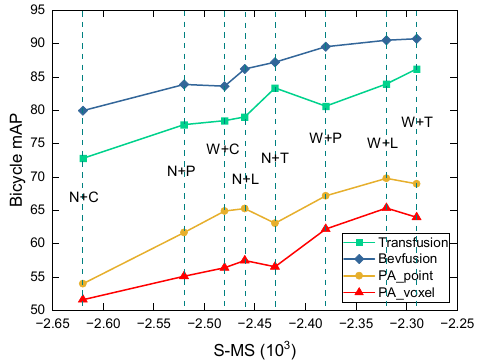}
\end{minipage}}
\begin{minipage}{2.3in}
\includegraphics[scale=0.75]{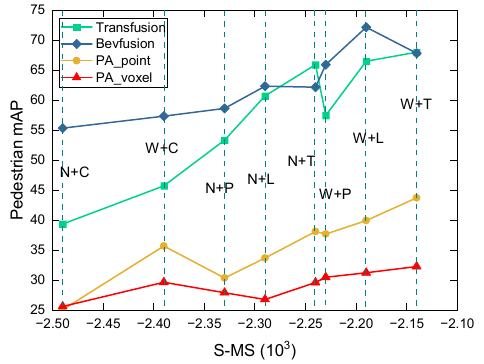}
\end{minipage}
\caption{The relationship between 3D detection mAP and unified surrogate metric (S-MS) under camera-LiDAR configurations, abbr. listed in Table \ref{tab1}}
\label{figure01}
\end{figure*}

\textbf{Influence of sensor configurations on 3D object detection.} In Fig. \ref{figure01} and Table \ref{table01}, we show the 3D target detection performance with different representative algorithms using different camera-LiDAR configurations.
The configurations of sensors drastically influence the detection performance, with a maximum fluctuation of 30\%. Specifically, for the effects of the camera and LiDAR configurations respectively, it is not difficult to find that for the same LiDAR configuration, the Wide camera configurations significantly outperform the Narrow camera configurations for up to 10\%; and for the same camera configuration, Trapezoid has the best detection performance for the majority of algorithms and objects. Under the Line LiDAR configuration, Pointaugmenting is the best for Bicycle while Bevfusion is the best for the Pedestrian, due to the point cloud detection characteristics of specific algorithms and the influence of POG for different objects.

\textbf{Correlation between the unified surrogate metric and detection performance.} Based on the proposed unified surrogate metric, we now analyze the interplay between 3D detection performance and camera-LiDAR configurations. Fig. \ref{figure01} shows the detection performance of Car, Bicycle, and Pedestrian at different S-MS values. While there are some fluctuations, the rising tendency for detection performance with increasing S-MS values is clear to see. The full configurations with abbreviations can be found in Table \ref{tab1}.

In addition to errors in data acquisition and stochastic model training, the fluctuations in the figure may be caused by two following factors. First, the same $\lambda$ value is adopted for all algorithms and target objects when calculating the S-MS values, but the real weight constant $\lambda$   may not be the same due to the differences in the characteristics of algorithms and target objects. Second, there exist some specific sensor configurations as preferences or attacks for specific learning-based algorithms \cite{kanbak2018geometric,engstrom2019exploring}. For instance, Line LiDAR configuration (L) has excellent detection performance for small objects (Bicycle and Pedestrian), as these objects are small in lateral dimension and Line configuration increases the point cloud density in the lateral direction. Under Pyramid LiDAR configuration (P), the performance of Transfusion to detect Car is drastically degraded due to the adversarial attack of Pyramid for Transfusion.

In addition, while the Wide camera configuration generally outperforms the Narrow camera configuration, detection performance with the Narrow + Trapezoid (N+T) configuration is sometimes better than with the Wide + other configurations, indicating that superior LiDAR sensor can compensate for camera deficiencies, and vice versa.

\textbf{Potential application analysis.} Our unified surrogate metric greatly accelerates the development, optimization, and evaluation of multi-sensor configurations for self-driving cars compared to the commonly seen R\&D process involving installing sensors, collecting data, training models, and evaluating performance. Our approach can efficiently evaluate different camera-LiDAR configurations by simply calculating the surrogate metric from the bounding boxes. Besides, the optimal solution of camera-LiDAR configurations for specific scenarios can be found based on our metrics.

\textbf{Future work.} The unified surrogate metric could be extended for more algorithms and include additional sensors\cite{hesailidar2023}, such as solid-state LiDAR and radar. Experiments on real sensors and the latest placements from the autonomous driving industry \cite{mercedesdrivepilot2023, motionaltech2023} are crucial for enhancing the effectiveness of our metric. The comparison of performance on data from simulation platforms and real-world data also represents an exciting frontier. Further exploration of optimizing the surrogate metric score to enhance the LiDAR placement would provide valuable insights.

\section{Conclusion}

In this paper, we investigate the influence of LiDAR and camera configurations on the performance of 3D object detection for autonomous driving. We propose a novel framework for evaluating LiDAR and camera configurations, including data acquisition, model training, and performance evaluation. We propose a unified surrogate metric that predicts 3D object detection performance for different camera and LiDAR configurations. We conduct extensive experiments with CARLA-collected data and representative camera-LiDAR fusion algorithms,  and the results have shown high consistency between our metrics and detection performance, providing new directions for the optimization of multi-sensor configuration in self-driving cars.

\bibliographystyle{IEEEtran}
\bibliography{root.bib}

\end{document}